\documentclass[10pt,twocolumn,letterpaper]{article}

\usepackage{cvpr}
\usepackage{times}
\usepackage{epsfig}
\usepackage{graphicx}
\graphicspath{{figs/}}
\usepackage{amsmath}
\usepackage{amssymb}
\usepackage{array}
\usepackage{comment}

\usepackage{xspace}
\makeatletter
\DeclareRobustCommand\onedot{\futurelet\@let@token\@onedot}
\def\@onedot{\ifx\@let@token.\else.\null\fi\xspace}
\def\eg{\emph{e.g}\onedot}

 \def\vs{\emph{vs}\onedot}
 
\def\etal{\emph{et al}\onedot}
\makeatother

\newcommand{\BigO}[1]{\ensuremath{\operatorname{O}\bigl(#1\bigr)}}

\usepackage{placeins}

\cvprfinalcopy 

\def\cvprPaperID{492} 

\ifcvprfinal\fi
\begin{document}

\title{Hierarchical Subquery Evaluation for Active Learning on a Graph}

\author{Oisin Mac Aodha \qquad Neill D.F. Campbell \qquad Jan Kautz \qquad Gabriel J. Brostow \\
University College London\\
{\small http://visual.cs.ucl.ac.uk/pubs/graphActiveLearning}
}


\maketitle

\begin{abstract}
To train good supervised and semi-supervised object classifiers, it is critical that we not waste the time of the human experts who are providing the training labels. Existing active learning strategies can have uneven performance, being efficient on some datasets but wasteful on others, or inconsistent just between runs on the same dataset. We propose perplexity based graph construction and a new hierarchical subquery evaluation algorithm to combat this variability, and to release the potential of Expected Error Reduction.

Under some specific circumstances, Expected Error Reduction has been one of the strongest-performing informativeness criteria for active learning. Until now, it has also been prohibitively costly to compute for sizeable datasets. We demonstrate our highly practical algorithm, comparing it to other active learning measures on classification datasets that vary in sparsity, dimensionality, and size. Our algorithm is consistent over multiple runs and achieves high accuracy, while querying the human expert for labels at a frequency that matches their desired time budget.

\end{abstract}

\section{Introduction}
Bespoke object recognizers are almost mature enough to be useful to people in practice. A major hurdle is how to procure enough training labels to tune a semi-supervised model for a specified classification task. While unskilled Mechanical Turkers are willing to label images of food at \$1.40 per image~\cite{noronha11:platemate}, the costs are massive for recruiting and paying specialists like doctors or scientists. 
Whether they are experts or part of an online crowd, people need practical and reliable Active Learning (AL) to suggest which unlabeled image they, as the oracle, should label next. Choosing the query images in the right order gives better classification after fewer interrogations of the oracle.

During a training session, the classifier model starts with only unlabeled examples, picks one, queries the human for its label, and then quickly re-trains the classifier so the process can repeat with queries selected among the remaining unlabeled examples. We therefore work within the popular graph based semi-supervised learning (SSL) framework, where each image is represented as a vertex in a weighted graph, weights encode similarity between image feature vectors, and vertices that have already been queried have labels. Whether the human is done providing class labels or not, classification of all datapoints is performed directly in feature space by propagating available label information over the graph. 

Designing a graph based AL framework requires three steps: 1) building a graph of the unlabeled datapoints in feature-space, 2) selection of an AL criterion for measuring the informativeness of possible queries, and 3) selecting an inference method for evaluating the criterion on the graph. There are many benefits to this framework, but forming the right combination of these three is an acknowledged challenge. The other steps are especially influenced by the AL criterion, chosen to decide which unlabeled image will be the next query. In particular, Expected Error Reduction (EER) is very attractive (see \S~\ref{sec:zhu_eer}), but naive incarnations of it are prohibitively costly. Each query put to the oracle is preceded by computing ``subqueries'' to each unlabeled example; a subquery simulates how the updated predictions \emph{would} change if that individual datapoint received this or that label from the oracle.

We therefore propose a method for graph construction that is good in its own right, but crucially, organizes the data so that the EER criterion can be exploited effectively. Building on our graph construction, our main contribution is the proposed hierarchical subquery evaluation, which allows us to ask the oracle for a label that maximizes EER, without having to compute EER exhaustively for all unlabeled images, and without heuristics that hurt the overall learning curve. Our many experiments show that the significant benefits of computing EER by traversing our hierarchical representation of the data are 1) that we can cope with datasets having a broad variety of sparsity, dimensionality, and size, 2) that we balance exploration \vs exploitation to get good accuracy quickly and refine decision boundaries as needed within the time budget specified by the user, and 3) that empirically, we have highly consistent accuracy when labeling a given dataset. Our experiments benchmark our approach against alternative AL criteria and alternative graph constructions, and establish the repeatability of our approach across different datasets.

\section{Related Work}
Here we cover only the most relevant related works, and recommend~\cite{Settles12activelearning} for a thorough overview of active learning. Active learning has been successfully applied to many different computer vision problems including tracking~\cite{vondrick2011video}, image categorization~\cite{joshi2009multi}, object detection~\cite{vijayanarasimhan2011large}, semantic segmentation~\cite{vezhnevets2012active}, and image~\cite{batra2010icoseg} and video segmentation~\cite{fathi2011combining}, with both human and automatic oracles~\cite{Karasev2014ActiveSelection}. Compared to the body of work on active learning in general, there are relatively few active learning methods for image classification which facilitate \emph{interactive} annotation. The challenge with creating interactive algorithms is that the time to retrain the model, once a labeled example is provided, can be long if not performed incrementally. This delay can also be further exacerbated by the type of active learning criterion used. Yao~\etal\cite{yao2012interactive} propose object detection based on efficient incremental training of Hough voting forests. Operating in real-time, their system is able to predict an annotation cost for an image and provides feedback to the user. However, they do not exploit the unlabeled data in the pool when updating their model. Batra~\etal\cite{batra2010icoseg} present a system for interactive image co-segmentation which asks the user to annotate the region deemed to be most informative by the current model. Wang~\etal\cite{wang2008active} perform cell image annotation using a semi-supervised graph labeling approach and exploit fast updating of the graph for interactive annotations. Unlike our work, they do not explore the merits of different active learning criteria.

\subsection{Semi-Supervised Active Learning}
In pool based active learning we have access to the unlabeled data up front, before querying the oracle. In contrast to standard supervised learning, semi-supervised learning (SSL) exploits structure in the unlabeled data. In this paper we are concerned with graph based SSL, however our proposed subquery evaluation scheme can be applied to any pool based active learning task where the unlabeled data is available during training. In graph based SSL, datapoints are represented as nodes in a graph and edges between the nodes encode similarity in feature space. The premise is that datapoints near each other in feature space share the same label. Graph based transductive algorithms can be efficient to evaluate in closed form, typically only requiring simple matrix operations to propagate label information around the graph.

\noindent\textbf{Graph Based SSL:}~%
Zhu~\etal\cite{zhu2003semi} propose an approach to SSL based on defining harmonic functions on Gaussian random fields. The advantage of their method is that, unlike graph cut based formulations~\cite{graphCutsBlum01}, it produces a probability distribution over the possible class labels for each datapoint. Having real probabilities opens the door to a broader range of active learning strategies. The LGC method of Zhou~\etal~\cite{zhou2004learning}, adds additional regularization by balancing the information a node receives from the labeled set and its neighbors, but at the expense of allowing a labeled node to change class. For both methods it is also possible to include a label regularization term to address class imbalance in the data~\cite{wang2008active}.

As the number of datapoints increases, it can quickly become infeasible to perform the large matrix inversions that are required by many graph based SSL algorithms. Iterative algorithms do not require a matrix inversion but can take many iterations to converge~\cite{Zhu2002labelprop, zhou2004learning}. Options to overcome this scalability issue include reducing the effective graph size using mixture models in feature space~\cite{zhu2005harmonic}, non-parametric regression of the labels through a subset of anchor nodes~\cite{liu2010large}, or assuming the data to be dimensionally separable in order to approximate the eigenvectors of the normalized graph Laplacian~\cite{fergus2009semi}. 

\noindent\textbf{Graph Construction:}~%
It is well known that graph based methods are highly sensitive to the choice of edge weights~\cite{jebara2009graph}. A standard approach for graph construction is to first sparsify the fully-connected graph and then reweight the remaining edges. Sparsification is important, because in higher dimensions, the distances between far away points become less meaningful. K-nearest neighbor and distance thresholding are common choices for sparsification. However, they suffer from the problem that the resulting graph can be uneven as there is no guarantee on the number of edges at each node. Approaches exist to guarantee regular graphs (the same number of edges at each node) but can be computationally costly \cite{jebara2009graph}. However, for a small decrease in graph quality, it is possible to build approximately regular graphs at reduced cost~\cite{wang2012fast}. In the reweighting step, a similarity measure between datapoints must be defined. One standard choice of similarity is the RBF kernel, and several methods have been proposed to define a suitable bandwidth parameter. If there are labeled datapoints it can be learned ~\cite{zhu2003semi}, alternatively it can be defined per dimension, based on the average distance between all neighbors~\cite{chapelle2006semi}, local distance~\cite{hein2006manifold}, or by direct optimization~\cite{wang2008label}. Wang~\etal\cite{wang2008graph} jointly learn the graph structure and label prediction by minimizing a cost function over the graph and its labeling. In this paper we propose a method for graph reweighting inspired by ideas from dimensionality reduction~\cite{hinton2002stochastic}. 


\noindent\textbf{Active Learning on Graphs:}~%
Many different active learning criteria exist in the literature. Methods range from random querying, uncertainty sampling, margin reduction, density sampling, expected model change, and expected error reduction~\cite{Settles12activelearning}. An optimal strategy would trade off between exploration and exploitation; initially exploring the space when there are few labels and uncertainty is high and then, when more annotations have been acquired, exploit this information to perform boundary refinement between the classes. Algorithms that switch between density based and uncertainty sampling typically require hyperparameters that are dataset specific~\cite{cebron2009active}, however more complex approaches strive to do this automatically~\cite{liadaptive,ebert2012ralf}. Expected error reduction (EER)~\cite{roy2001} performs this trade off naturally. Instead of measuring a surrogate, it seeks out datapoints that will make the overall class distributions on the unlabeled data more discriminative by attempting to reduce the model's future generalization error.

However, full EER requires $\BigO{N^2}$ operations to determine which example minimizes the expected error under the current model, where $N$ is the size of the dataset. This complexity stems from needing to retrain the model for each of the $N$ subqueries in the unlabeled pool to evaluate their expected error. Efficient update methods for some commonly known algorithms exist, \eg in graph based SSL making full EER only feasible on small graphs. Zhu~\etal\cite{Zhu03combiningactive} demonstrated the superior performance of EER over other active learning criteria when combining it with their Gaussian fields formulation~\cite{zhu2003semi}, and this serves as one of our baselines. 


\noindent\textbf{Clustering Approaches:}~%
To cope with larger datasets, different approaches have been proposed to reduce the number of subqueries that must be evaluated. Strategies include 
only considering a subsample of the full data~\cite{roy2001}, or using the inherent structure of the data to limit influence and selection of subqueries~\cite{kddActiveMethods}. Using the same manifold assumption as SSL, these 
methods cluster the data in its original feature space. Macskassy~\cite{kddActiveMethods} explores graph based metrics, commonly used in community detection, to 
identify cluster centers (each assumed to contain the same class) that are then evaluated using EER. 
This is related to the hierarchical clustering method for category discovery of Vatturi~\cite{vatturi2009category}. However, by limiting subqueries to cluster centers, these clustering based approaches are unable to perform boundary refinement. 

The hierarchical clustering, in~\cite{dasgupta2008hierarchical}, is used to define bounds on sampling statistics. Every one of their samples (a full query to the oracle) is randomly selected from a strict partition of a prespecified clustering (similar to a breadth first search) and only shares label information within its cluster. 
Our proposed method also uses a hierarchical representation, but differs as it uses the hierarchy for efficient sampling using EER, with the added advantages of graph based SSL, without sacrificing the ability to refine class boundaries.

\newcommand{\Prob}   {\ensuremath{P\xspace}}
\newcommand{\LabelledData} {\ensuremath{\mathcal{D}_l}}
\newcommand{\UnlabelledData} {\ensuremath{\mathcal{D}_u}}
\newcommand{\EstOutProb} {\ensuremath{\hat{P}_{\LabelledData}\xspace}}
\newcommand{\Error}   {\ensuremath{\mathcal{E}}}
\newcommand{\Loss}   {\ensuremath{L\xspace}}
\newcommand{\Indicator}  {\ensuremath{\mathrm{I}}}
\newcommand{\vx}   {\ensuremath{\mathbf{x}}}
\newcommand{\PlusQuery}   {\ensuremath{+ \!(\vx_q,y_q)}}
\newcommand{\PlusQueryEqyPrime}   {\ensuremath{+ \!(\vx_q,y_q = y')}}
\newcommand{\Expect}  {\ensuremath{\mathrm{E}\xspace}}

\section{Graph Based Semi-Supervised Framework}
Here we review graph based SSL, and detail our innovations in \S~\ref{sec:OurApproach}. In pool based learning, one has a dataset $\mathcal{D} = \left\{(\mathbf{x}_1, y_1), ..., (\mathbf{x}_N, y_N)\right\}$ where each $\mathbf{x}_i$ is a $Q$ dimensional feature vector and $y_i\in {1, ..., C}$ is its corresponding class label. 
We split $\mathcal{D}$ into two disjoint sets $\UnlabelledData$ and $\LabelledData$, corresponding to the sets of unlabeled and labeled examples. For active learning, the set of labeled examples is initially empty as only the oracle knows the values of each $y_i$. One can define a graph $G$ with a set of vertices $\mathcal{V}$, 
corresponding to the pool of $N$ examples in $\mathcal{D}$, and the set of edges is represented by a connectivity weight matrix $W \in \mathbb{R}^{N\times{}N}$. Each entry $w_{ij}$ in $W$ represents the similarity in some feature space between datapoints $\mathbf{x}_i$ and $\mathbf{x}_j$. Our goal is to estimate the distribution over the class labels for each of the nodes in the graph, $f_{ic} = P(y_i \!=\! c\,|\,\mathbf{x}_i)$. In matrix notation, these distributions, $F$, are represented as an $N\times C$ matrix, where each row is a different datapoint.

Zhu~\etal\cite{zhu2003semi} propose a method for semi-supervised learning based on Gaussian random fields and harmonic energy minimization (GRF). Their harmonic energy minimization can be computed in closed form using matrix operations on the graph Laplacian,
\begin{equation}
F_u = (D_{uu} - W_{uu})^{-1}W_{ul}Y_{l},
\end{equation}where $D$ is a diagonal matrix with entries $d_{ii} = \sum_j w_{ij}$. The matrices are split into labeled and unlabeled parts
\begin{equation}
 W = \left[ \begin{array}{cc} W_{ll} & W_{lu}\\ W_{ul} & W_{uu}\end{array} \right],\mbox{ ~and }
Y = \left[ \begin{array}{c} Y_l \\ Y_u \end{array} \right].
\end{equation}
Again using matrix notation, $Y$ is the same size as $F$ where all entries are set to $0$ except where the oracle labels datapoint $\mathbf{x}_{i}$ with class $c$ making  $y_{ic} = 1$.


\subsection{Expected Error Reduction \label{sec:zhu_eer}}

Let $\Prob(y|\vx)$ be the unknown conditional distribution of output $y$ over input $\vx$, and $\Prob(\vx)$ be the marginal input distribution.  Taking the labeled data $\LabelledData$, we can produce a learner that estimates the class output distribution $\EstOutProb(y|\vx)$ for a given input $\vx$. The expected error of such a learner is
\begin{equation}\label{eqEE1}
\Error_{\EstOutProb} = \int_{\vx} \Loss\left( \Prob(y|\vx), \EstOutProb(y|\vx) \right) \,,
\end{equation}
where we define $\Loss(\cdot,\cdot)$ as a loss function that quantifies any error between the predicted output and the true value. In our learning problem, we consider multi-class classification tasks and use a 0/1 loss function
\begin{equation}
\Loss\left( \Prob(y|\vx), \EstOutProb(y|\vx) \right) = \sum_{y = 1}^{C} \Prob(y|\vx) \; \Indicator\left[ y \neq \hat{y} \right] \,,
\end{equation}
where $\hat{y} = \arg\max_{y} \EstOutProb(y | \vx)$ is the learner's MAP estimate of the class of $\vx$, and $\Indicator[\cdot]$ is a binary indicator function.

In the case of graph based SSL, we represent the marginal input distribution by the set of input samples $\{\vx_i\}$ and evaluate the integral of (\ref{eqEE1}) as a summation over this set to produce
\begin{equation}
\Error_{\EstOutProb} = \sum_{i=1}^{N} \sum_{y_i = 1}^{C} \Prob(y_i|\vx_i) \; \Indicator\left[ y_i \neq \hat{y_i} \right]
\end{equation}
as the expected error.  In practice, the true conditional distribution $\Prob(y|x)$ is unknown, so we approximate it using the current estimate of the learner $\EstOutProb(y|x)$.

In the context of active learning, we would like to select the oracle's next query $(\hat{\vx}_q,\hat{y}_q)$ from the unlabeled data $\UnlabelledData$, such that adding it to the labeled data $\LabelledData$ would result in a new learner with a lower expected error. This leads to a greedy selection strategy. First, we determine the expected error (or \emph{risk}) for combinations of each unlabeled example $\vx_q \in \UnlabelledData$ taking each possible label $y_q \in \{1 .. C\}$
\begin{equation}
\Error_{\EstOutProb}^{\PlusQuery} = \sum_{i=1}^{N} \sum_{y_i = 1}^{C} \EstOutProb^{\PlusQuery}(y_i|\vx_i) \; \Indicator\!\left[ y_i \neq \hat{y_i}^{\PlusQuery} \right] \,,
\end{equation}
where $\EstOutProb^{\PlusQuery}$ is the learner with $(\vx_q,y_q)$ added to the labeled data. We then calculate the expectation of this risk across the possible label values for $y_q$. We use the learner's current posterior $\EstOutProb(y_q|x_q)$ to approximate the unknown true distribution across $y_q$ to provide
\begin{equation}
\Expect \left[ \Error_{\EstOutProb}^{\PlusQuery} \right] = \sum_{y' = 1}^{C} \EstOutProb(y_q \!=\! y'\,|\,\vx_q) \;\Error_{\EstOutProb}^{\PlusQueryEqyPrime}
\end{equation}
as the expected risk. Finally, we select the query $\hat{\vx}_q$ with the smallest expected risk.  For the remainder of the paper, we refer to this expected risk as the \emph{expected error} that the EER criterion seeks to minimize.

Zhu~\etal\cite{Zhu03combiningactive} integrated active learning into their GRF framework by exhaustively calculating the expected error over all possible unlabeled nodes. Even with the proposed matrix update efficiencies of Zhu~\etal, calculating the expected error for a datapoint is a linear operation and evaluating it over all unlabeled examples results in a time complexity of $\BigO{{|\UnlabelledData|}^2}$. This quadratic cost is prohibitively expensive as the dataset increases in size. We address this limitation using our proposed hierarchical subquery sampling approach presented in \S~\ref{sec:hierarchical_approach}.

\section{Hierarchical Subquery Evaluation}\label{sec:OurApproach}

Our method uses the EER active learning criterion while overcoming the expense of exhaustive sampling. It does this without sacrificing the desirable exploration/exploitation properties of EER, an issue with previous subsampling approaches. Before we discuss our hierarchical subquery search method, we first describe our graph construction technique that we have found to work well with the EER criterion and to be robust across a wide variety of datasets.

\subsection{Perplexity Based Graph Construction}
As noted previously, graph based SSL algorithms are very sensitive to the choice of similarity matrix $W$. If two datapoints $\vx_i$ and $\vx_j$ have the same label, we want their corresponding affinity $w_{ij}$ to be high, and if they are different we want it to be low. One popular choice of similarity kernel is the radial basis function (RBF), 
\begin{equation}
w_{ij} = \exp(-\gamma{}_i\lVert \mathbf{x}_i - \mathbf{x}_j \rVert{}_2^2). 
\end{equation} Here we use the $L_2$ distance, but other distances may be more appropriate depending on the data representation (\eg histograms). We have now introduced a set of parameters $\gamma_i$ that control the bandwidth of the kernel. A single choice of $\gamma$ is unlikely to be optimal across the whole dataset. We want each $\gamma_i$ to model the density of local space. Intuitively, we want a larger value of $\gamma_i$ in dense regions of the feature space and a smaller value in more sparse regions. We now define our similarity based on a successful unsupervised technique from dimensionality reduction.

In Stochastic Neighbor Embedding (SNE)~\cite{hinton2002stochastic} the non-symmetric similarity between points is represented as a conditional probability. $w_{ji}$ can be interpreted as the probability that $\mathbf{x}_i$ would pick $\mathbf{x}_j$ as its neighbor assuming there is a Gaussian with variance $\sigma_i^2$ centered at $\mathbf{x}_i$,  where $\gamma_i = 1/(2\sigma_i^2)$. We perform the same binary search as SNE to find the values of $\gamma_i$ that best match a given level of \emph{perplexity} (a measure of the effective number of local neighbors). The perplexity for a given choice of $\gamma_i$ is defined as
\begin{equation} 
\mathrm{Perp}(\gamma_i) = 2^{ -\sum_j w_{ji} \log_2 w_{ji}} \,.
\end{equation} 
We enforce a valid similarity matrix $W$ by symmetrizing the conditional probabilities, so $w_{ji} = \frac{1}{2}(w_{ij} + w_{ji})$. 

%
%
%
\subsection{Hierarchical EER \label{sec:hierarchical_approach}}
The EER criterion dictates that we pick the datapoint giving the lowest expected error to be labeled next. We refer to calculating the expected error of a single unlabeled datapoint as a subquery; the complexity of a single subquery is linear in the number of unlabeled datapoints. Together, the subqueries are internal calculations used to determine the next query that is sent to the oracle for labeling. We want to find the next query within a specified query budget.  This means we do not have sufficient time to perform subqueries on all possible unlabeled nodes since this results in a quadratic cost (\S~\ref{sec:zhu_eer}). Instead, we must identify an adaptive number of the best subqueries to sample within an allotted time, ideally sub-linear in the number of unlabeled nodes. 

\begin{figure*}[t!]
\def\imagetop#1{\vtop{\null\hbox{#1}}}
\centering
\setlength{\tabcolsep}{5pt}
\begin{tabular}{cc}
\includegraphics[height=0.26\linewidth]{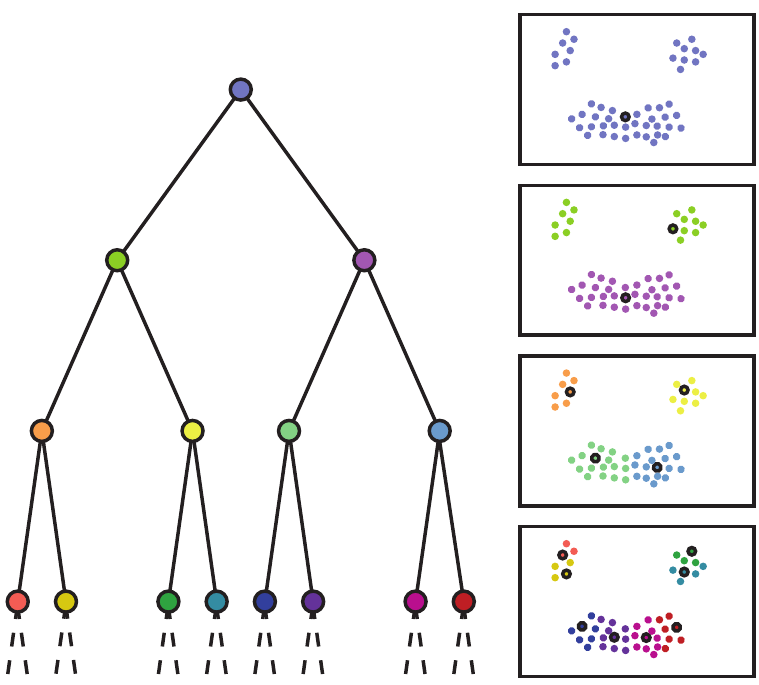} &
\includegraphics[height=0.26\linewidth]{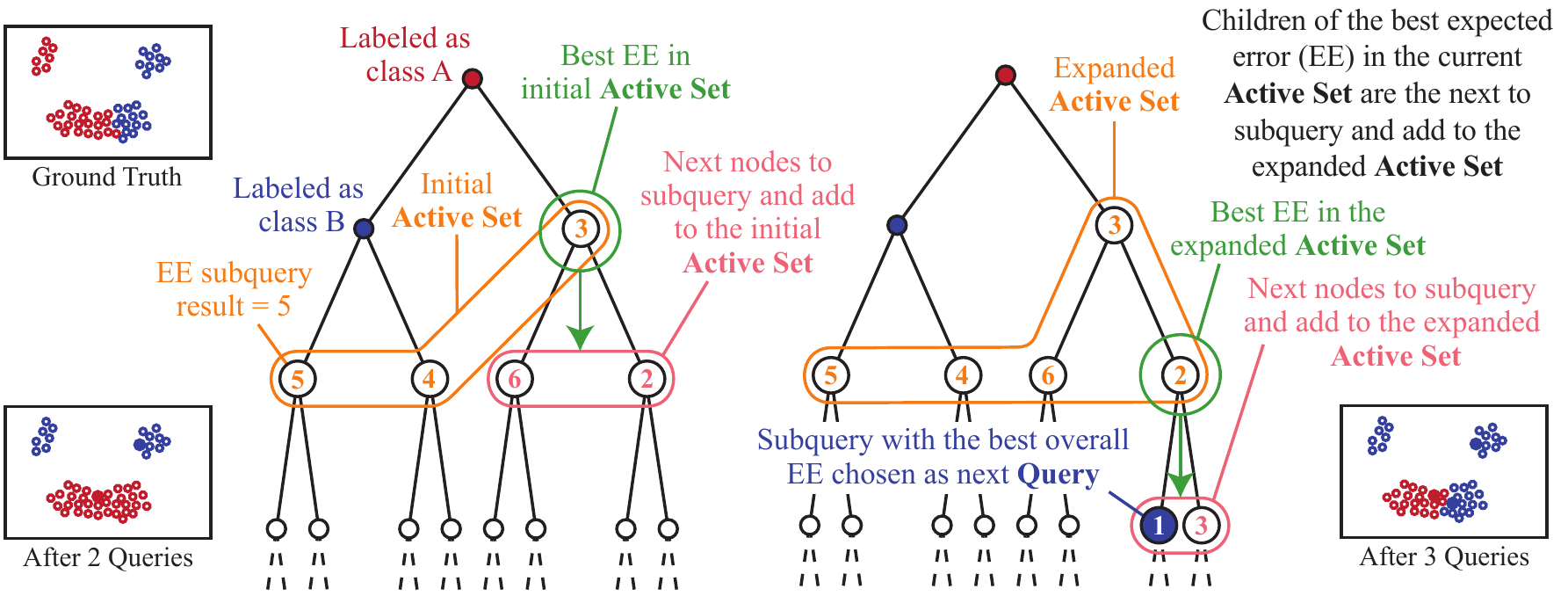} \\
\small{\textbf{(a)} The AuthorityShift clustering} & \small{\textbf{(b)} Our hierarchical subquery sampling algorithm for the next active learning query}
\end{tabular}
\caption{Hierarchical clustering and subquery sampling strategy. 
%
(a)~A hierarchical clustering is built using~\cite{cho2010authority}, shown here as a tree. 
At each level, every node in the tree is represented by a unique allocation (denoted by color) to a specific datapoint (the \emph{authority point} in bold). 
(b)~We use a hierarchical algorithm to determine the subqueries to perform; a subquery evaluates the expected error (EER criterion) shown as a number inside the node. 
(left)~An \emph{active set}, shown in orange, is constructed containing the children of labeled nodes; these are evaluated as the first subqueries, prioritizing from top to bottom. The active set is then expanded in a greedy fashion by including the children of the subquery with the lowest expected error, shown in pink. (right)~We repeat this process until we have exhausted our subquery budget. The query for the oracle to label is chosen as the subquery with the lowest expected error (greatest EER). }
\label{clustSampStrat}
\end{figure*}

The smooth nature of the harmonic solutions, with respect to proximity of nodes on the graph, creates a redundancy in densely sampling all nodes; neighboring nodes will likely produce a very similar reduction in error when labeled. A hierarchical clustering of the graph, for example Figure~\ref{clustSampStrat}(a), exploits these local correlations between neighboring nodes. 
Previous approaches to reducing the number of subqueries have included random sub-sampling~\cite{roy2001} and using community detection to propose candidates~\cite{kddActiveMethods}. The latter method is equivalent to performing a breadth first (coarse to fine) search of a cluster hierarchy where graph communities are represented as high level clusters. Similar breadth first searches of hierarchies have been used in active learning, albeit without the EER criterion~\cite{dasgupta2008hierarchical,vatturi2009category}.

The main advantage of the EER criterion is that it will trade-off the reduction in error achieved by either labeling an unknown region (exploration) or refining the decision boundaries under its current model (exploitation). Typically, the exploration mode will label nodes high up in the hierarchy whereas the detailed boundary refinement will occur in the leaves of the tree. While a breadth first approach can achieve good initial results, the active learner is stuck in an exploratory mode since it is effectively sampling on a graph density measure.

In our proposed approach, we allow the EER measure to perform the exploration/exploitation trade-off while still sub-sampling the unknown nodes to dramatically reduce the number of subqueries and therefore the cost. We achieve this by performing an adaptive search of the hierarchy.



\subsection{Hierarchical Subquery Sampling}

\noindent\textbf{Authority-Shift Hierarchy Creation:}~%
We provide an illustrative example of the hierarchical clustering in Figure~\ref{clustSampStrat}(a). We make use of the Authority-Shift algorithm of Cho and Lee~\cite{cho2010authority}. It does not require a feature space but operates on the perplexity graph directly. This technique produces a hierarchical clustering on a graph by authority seeking: the process of allocating each node to a local `authority' node (that represents the cluster). The calculation explores the steady state of a set of random walks on the graph at an appropriate scale. By increasing the scale parameter iteratively, a hierarchy of clusters can be built up to form a tree. This approach has two advantages. First, each cluster in the tree is represented by a specific datapoint that can be used to perform a subquery. Second, the clusters themselves encode walks on the graph under the same transition matrix used to evaluate the harmonic function, and therefore produce a summary of the results of calculating the expected error for all the datapoints in the cluster.



\noindent\textbf{Subquery Sampling:}~%
An overview of our hierarchical sampling algorithm is provided in Figure~\ref{clustSampStrat}(b). We differ from previous breadth first searching strategies by allowing an adaptive search on the tree to greedily seek for the minimum reduction in expected error. Referring to the diagram, consider a set of data with the cluster hierarchy of Figure~\ref{clustSampStrat}(a), where two nodes have already been queried and labeled; see the left side of Figure~\ref{clustSampStrat}(b). 
First, we build an \emph{active set} of unlabeled nodes containing the children of labeled nodes, starting at the root. We proceed to perform a batch of subqueries of this active set (shown in orange) to obtain the expected error (the numbers inside the nodes). We then expand the active set by adding the children of the subquery in the current active set with the minimum expected error (shown in pink). As the children are added to the active set, they are evaluated as subqueries; see the right side of Figure~\ref{clustSampStrat}(b). This process repeats until we have exhausted our budget of subqueries (a limit on the size of the active set). We now select the member of this active set with the minimal expected error as the next query to be labeled by the oracle. 
We prioritize the subquery evaluation by the level in the hierarchy (top-to-bottom) and then by ranking the nodes based on the total number of their descendants. 

%
%
%
%
\begin{table*}[ht] 
 \setlength\tabcolsep{5pt}
 \centering 
 \small
 \scalebox{0.9}{%
 {\setlength{\extrarowheight}{3pt}
 \begin{tabular}{l|cccc|cccc|c|ccc} 
 \hline 
Dataset  & N & D & C & Feat & \phantom{M}rand\phantom{M} & margin & entropy & RALF~\cite{ebert2012ralf} & Zhu~\cite{Zhu03combiningactive} & randS~\cite{roy2001} & bFirst~\cite{kddActiveMethods} & \bf{HSE (ours)} \\ \hline
Glass~\cite{uciData} & 214 & 10 & 6 & - & 0.732 & 0.605 & 0.599 & 0.763 & \textbf{0.818} & 0.810 & 0.782 & 0.804 \\ 
Ecoli~\cite{uciData} & 336 & 7 & 8 & - & 0.759 & 0.781 & 0.788 & 0.812 & 0.832 & 0.829 & 0.782 & \textbf{0.833} \\ 
Segment~\cite{uciData} & 635 & 18 & 7 & - & 0.811 & 0.717 & 0.680 & 0.832 & \textbf{0.903} & 0.896 & 0.840 & 0.896 \\ 
FlickrMat~\cite{flickrmaterial_paper} & 1000 & 50 & 10 & {\footnotesize PCA BoW} & 0.172 & 0.131 & 0.125 & 0.242 & \textbf{0.261} & 0.244 & 0.249 & 0.259 \\ 
Coil20~\cite{nene1996columbia} & 1440 & 20 & 20 & {\footnotesize PCA} & 0.558 & 0.392 & 0.456 & 0.713 & 0.729 & 0.757 & 0.756 & \textbf{0.760} \\ 
LFW10~\cite{LFW_paper} & 1456 & 50 & 10 & {\footnotesize PCA BoW} & 0.310 & 0.261 & 0.247 & 0.352 & 0.421 & 0.419 & 0.410 & \textbf{0.422} \\ 
UIUCSport~\cite{UIUC_paper} & 1579 & 50 & 8 & {\footnotesize PCA BoW} & 0.425 & 0.405 & 0.300 & 0.604 & 0.650 & 0.669 & 0.624 & \textbf{0.671} \\ 
Gait~\cite{hospedales2013finding} & 2353 & 25 & 9 & {\footnotesize PCA} & 0.506 & 0.434 & 0.313 & 0.650 & 0.668 & 0.665 & 0.669 & \textbf{0.696} \\ 
Oil~\cite{uciData} & 3000 & 12 & 3 & - & 0.927 & 0.800 & 0.798 & 0.916 & 0.943 & 0.948 & 0.979 & \textbf{0.986} \\ 
Caltech4~\cite{Fei-fei06one-shotlearning} & 3188 & 20 & 4 & {\footnotesize PCA BoW} & 0.953 & 0.922 & 0.936 & 0.966 & 0.986 & 0.988 & \textbf{0.993} & 0.990 \\ 
Eth80~\cite{ebert2012ralf} & 3280 & 576 & 8 & {\footnotesize HoG} & 0.531 & 0.359 & 0.370 & 0.660 & 0.649 & 0.603 & 0.665 & \textbf{0.675} \\ 
CpPascal08~\cite{ebert2012ralf} & 4450 & 576 & 20 & {\footnotesize HoG} & 0.091 & 0.075 & 0.079 & \textbf{0.277} & 0.074 & 0.073 & 0.167 & 0.184 \\ 
15Scenes~\cite{15_Scenes} & 4485 & 50 & 15 & {\footnotesize PCA BoW} & 0.255 & 0.236 & 0.144 & 0.548 & 0.535 & 0.505 & 0.469 & \textbf{0.573} \\ 
\hline 
Mean  &  &  &  &  & 0.541 & 0.471 & 0.449 & 0.641 & 0.651 & 0.647 & 0.645 & \textbf{0.673} \\ 
Wins  &  &  &  &  & 0 & 0 & 0 & 1 & 3 & 0 & 1 & \textbf{8} \\ \hline 
 
 \end{tabular}}%
 }
 \vspace{6pt}
 \caption{Datasets used for our evaluation where N, D, C, and Feat refer to the number of datapoints, dimensionality, number of classes, and representation. Results are presented as areas under the learning curve (1.0 is ideal). The learning curves for a subset of these datasets are depicted in Figure~\ref{graph_compar_fig}. Our method outperforms the other baselines, including full EER~\cite{Zhu03combiningactive} despite requiring far fewer subquery evaluations.}
 \label{mainResultsTab}
\end{table*}
%
%
\begin{figure*}[ht]
\centering
		\includegraphics[width=1.0\linewidth]{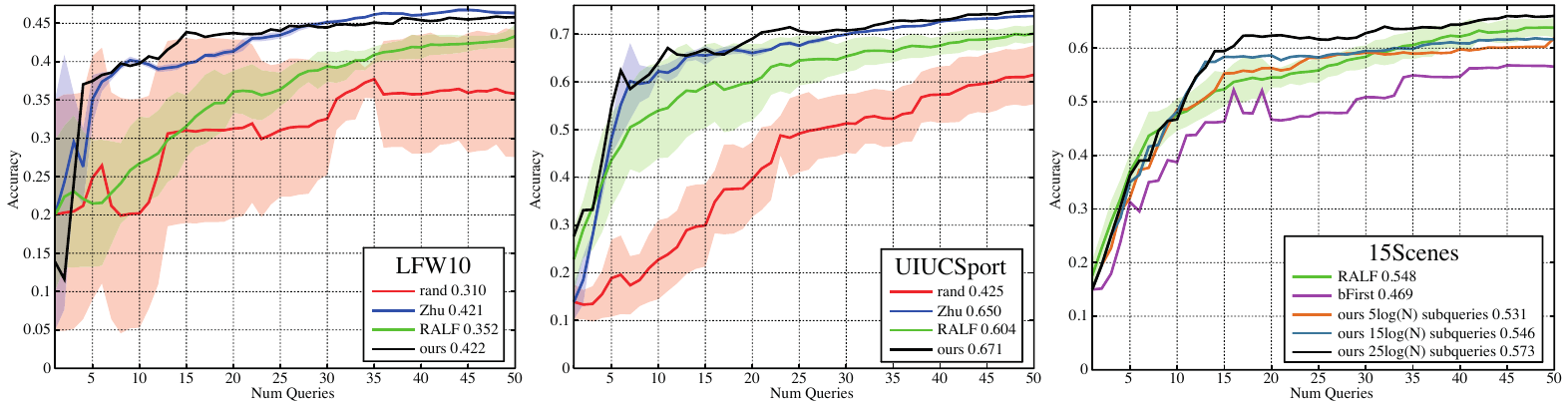}
\caption{Learning curves illustrating the performance of our approach versus three other baselines from Table~\ref{mainResultsTab}. The shaded regions around each learning curve represents one standard deviation. Our method gives superior results compared to that of Zhu~\etal\cite{Zhu03combiningactive} and as it is deterministic, results do not vary over different runs. In the last plot we illustrate the effect of increasing the number of subqueries for our method. As the number increases, so does the area under the curve.}
\label{graph_compar_fig}
\end{figure*}

The boxes in Figure~\ref{clustSampStrat}(b) provide a toy illustration of the advantage of this approach. To refine the boundary between the two classes, we need to ask the oracle to label nodes at the edges of clusters; these are usually found low down in the hierarchy. Because the EER improves as one moves toward a decision boundary, the active set can move down into the tree when the EER criterion favors exploitation over the  improvement of exploration; exploration occurs by labeling clusters at the top of the tree. Under breadth first search, a large number of queries would have to be performed before reaching nodes at the exploitation depth. As the learning curve evolves, the boundary refinement nodes will become increasingly localized, making it more unlikely that they will be found by random subqueries alone. We always take the root node of the tree as our first query (an open question for many algorithms) which we observe empirically to confer good performance and makes our algorithm deterministic. The tree construction means that the entire hierarchy has the potential to be navigated in $\BigO{N log(N)}$.

\vspace{-2pt}
\section{Experiments}
Table~\ref{mainResultsTab} describes the $13$ vision and standard machine learning datasets used for our experiments. These were chosen because they vary in size, density in their respective feature spaces, and have different numbers of classes. For all experiments, we start out with $3$ random queries, construct graphs with $10$ nearest neighbors based on the $L_2$ distance, use a perplexity value of $30$, and query the oracle $50$ times. For our method (HSE), we set the number of subqueries to be $25\log(N)$, where $N$ is the number of datapoints for a given dataset, and the initial queries are set as the first $3$ nodes in the hierarchy. Data and code are available on our project webpage.

\noindent\textbf{Graph Construction:}~%
Graph based SSL algorithms can produce inferior performance with poor graphs. Using the method of Zhu~\etal\cite{Zhu03combiningactive} to evaluate graphs, Table~\ref{graphcomparTable} compares our perplexity based graph construction method to four other baseline algorithms, testing this contribution in isolation. For {\it mean}, the bandwidth of the RBF kernel is set using the average distance between neighbors. For {\it binary}, we set a constant value for any two nodes that are connected and zero elsewhere. For {\it knn}, the bandwidth is set per datapoint proportional to its K-nearest neighbors. Finally, {\it lle} is the local linear embedding approach of~\cite{wang2008label}. Our perplexity based graph performs best overall. 

\begin{table}[t] 
 \centering 
 \small
 \setlength\tabcolsep{5pt}
 \scalebox{0.9}{%
 {\setlength{\extrarowheight}{3pt}
 \begin{tabular}{l|ccccc} 
 \hline 
Dataset  & mean & binary & knn \cite{hein2006manifold} & lle \cite{wang2008label} & \bf{per (ours)} \\ \hline 
Glass & 0.775 & 0.743 & 0.758 & 0.787 & \textbf{0.818} \\ 
Ecoli & 0.795 & 0.768 & 0.777 & 0.791 & \textbf{0.832} \\ 
Segment & 0.837 & 0.860 & 0.853 & 0.892 & \textbf{0.903} \\ 
FlickrMat & 0.196 & 0.159 & 0.198 & 0.222 & \textbf{0.261} \\ 
Coil20 & 0.641 & 0.597 & 0.616 & \textbf{0.729} & \textbf{0.729} \\ 
LFW10 & 0.362 & 0.356 & 0.365 & 0.381 & \textbf{0.421} \\ 
UIUCSport & 0.528 & 0.452 & 0.527 & 0.529 & \textbf{0.650} \\ 
Gait & \textbf{0.686} & 0.646 & 0.672 & 0.579 & 0.668 \\ 
Oil & 0.941 & 0.937 & 0.924 & \textbf{0.962} & 0.943 \\ 
Caltech4 & 0.981 & 0.973 & 0.977 & 0.971 & \textbf{0.986} \\ 
Eth80 & 0.572 & 0.596 & 0.562 & 0.604 & \textbf{0.649} \\ 
CpPascal08 & 0.146 & 0.102 & \textbf{0.159} & 0.141 & 0.074 \\ 
15Scenes & 0.344 & 0.304 & 0.353 & 0.378 & \textbf{0.535} \\
\hline 
Mean  & 0.600 & 0.576 & 0.595 & 0.613 & \textbf{0.651} \\ 
Wins  & 1 & 0 & 1 & 2 & \textbf{10} \\ \hline 
 \end{tabular}}%
 }
 \vspace{6pt}
 \caption{Comparison of different graph construction methods. The results represent area under learning curves for the GRF method of Zhu~\etal\cite{Zhu03combiningactive}. Our perplexity based method outperforms the other baselines.}
 \label{graphcomparTable}
\end{table}

\noindent\textbf{Active Learning Criteria:}~%
We compare our algorithm to seven different baselines, including GRF~\cite{zhu2003semi} with random, entropy, and margin based criteria~\cite{Settles12activelearning}, full EER~\cite{Zhu03combiningactive}, and the recent time varying combination approach RALF~\cite{ebert2012ralf}. We also compare to two different subquery evaluation strategies, random~\cite{roy2001} and breadth first~\cite{kddActiveMethods}. Both competing subquery strategies are evaluated using the same number of subqueries as our method. All methods use our perplexity based graph with the exception of RALF which uses a binary based graph representation. Empirically, we found RALF to perform worse using other graphs. Table~\ref{mainResultsTab} summarizes our overall results as area under the learning curve on the unlabelled set. 

Interestingly, our method outperforms full EER which requires $\BigO{N^2}$ computations. We note that the full EER is still a greedy algorithm at each iteration and therefore, not necessarily globally optimal. Our approach will encourage exploration at the start, when only a few queries have been performed and the active set is at the top of the hierarchy, which is observed to offer improved performance.

One noticeable exception is the Cropped Pascal dataset from~\cite{ebert2012ralf}. Due to the high variability in each class, it is likely that this dataset does not conform to the clustering assumption of semi-supervised learning. Using an iterative label propagation algorithm with few propagation steps prevents RALF~\cite{ebert2012ralf} from overfitting the dataset at the expense of worse marginals. Figure~\ref{graph_compar_fig} illustrates learning curves for a subset of the datasets. 

\begin{table}[t] 
 \centering 
 \small 
 \scalebox{0.9}{%
 {\setlength{\extrarowheight}{3pt} 
 \begin{tabular}{l|ccc} 
 \hline 
Dataset  & RALF~\cite{ebert2012ralf} & Zhu~\cite{Zhu03combiningactive} & \textbf{HSE (ours)} \\ \hline
Glass & 0.003 & 0.008 & 0.291 \\ 
Ecoli & 0.004 & 0.016 & 0.302 \\ 
Segment & 0.005 & 0.056 & 0.276 \\ 
FlickrMat & 0.007 & 0.231 & 0.136 \\ 
Coil20 & 0.011 & 0.950 & 0.369 \\ 
LFW10 & 0.009 & 0.535 & 0.172 \\ 
UIUCSport & 0.009 & 0.507 & 0.172 \\ 
Gait & 0.012 & 1.610 & 0.257 \\ 
Oil & 0.010 & 1.008 & 0.339 \\ 
Caltech4 & 0.011 & 1.435 & 0.351 \\ 
Eth80 & 0.014 & 2.793 & 0.378 \\ 
CpPascal08 & 0.041 & 12.189 & 0.753 \\ 
15Scenes & 0.033 & 9.405 & 0.710 \\ 
\hline 
 \end{tabular}}%
}
\vspace{6pt}
 \caption{Average time (in seconds) per query for active learning methods with different area-under-learning curve and across datasets of varying complexity. Both RALF and HSE pick the next query in under a second. In our method, we allow $25\log(N)$ subqueries per query rather than the full $N^2$ required for the Zhu method.}
\label{timingResultsTab}
\end{table}

Table~\ref{timingResultsTab} depicts the average time required to present the next query to the user for the different active learning methods. RALF~\cite{ebert2012ralf} scales linearly while full EER~\cite{Zhu03combiningactive} soon becomes impractical as the the number of examples increases. On average, our method computes queries in under a second and performs better than both methods in terms of accuracy.

%
%

%
%

\section{Discussion}
Accurate AL is the key to saving human effort, but speed is also a factor when a human oracle's patience is finite. Generalizing slightly, our Active Learning approach performs as accurately or better than Zhu~\etal\cite{Zhu03combiningactive}, but does so with an effective computational complexity on par with Ebert~\etal\cite{ebert2012ralf}. Their computational complexities are $\BigO{N^2}$ and $\BigO{N}$ respectively, while ours is $\BigO{N\log(N)}$ with a low $\log(N)$. In practice, with our Matlab implementation and default settings (used throughout), the combined subqueries needed to pick the oracle's next query finished in under a second, even for the largest datasets tested. For bigger datasets, users may opt to use our algorithm with fewer subqueries to keep the labeling interactive.
Both those main competitors are very good, excelling on specific datasets. Therefore it is important that validation of our AL approach has considered accuracy, efficiency, and generalizability to a variety of situations. The online supplementary material further illustrates that across these datasets, our hierarchical subquery evaluation leads to accurate results in the form of steep learning curves with large areas under the curve, and that these results are consistent across multiple runs, as plotted with $\pm1$ standard deviation from each curve's mean.

To tease apart the impact of our hierarchical subquery evaluation \vs our perplexity-based graph construction, we gave our graphs to the compatible AL baseline algorithms. Zhu~\etal is among them, and without our graphs, performs worse than RALF. Within the flexible graph based SSL framework, other choices can also have an impact, so as part of the supplemental files, we also show that LGC, used by RALF, is not as effective for our label propagation as Zhu~\etal's GRF.

There are several exciting avenues for future work. Our approach is transductive, so it would be attractive to either embed new datapoints into our existing graph online, or to transfer learned parameters to an inductive model. It would also be interesting to budget subqueries to account for some labels taking more of the oracle's time or effort than others. Finally, our similarity graph is computed once offline and never updated. In future, we may wish to use the label information from the user to learn a feature representation online.


\noindent\textbf{Acknowledgements:}~%
Funding for this research was provided by EPSRC grants EP/K015664/1, EP/J021458/1 and EP/I031170/1.

{\footnotesize
\bibliographystyle{ieee}
\bibliography{graphactivelearningbib}
}

\end{document}